\documentclass[letterpaper, 10 pt, conference]{ieeeconf}  % Comment this line out if you need a4paper

\IEEEoverridecommandlockouts                              % This command is only needed if 
\usepackage{graphics} % for pdf, bitmapped graphics files
\usepackage{epsfig} % for postscript graphics files
\usepackage{mathptmx} % assumes new font selection scheme installed
\usepackage{times} % assumes new font selection scheme installed
\usepackage{amsmath} % assumes amsmath package installed
\usepackage{amssymb}  % assumes amsmath package installed
\usepackage{booktabs}

\usepackage[font=footnotesize]{caption}
\usepackage{titlesec}
\usepackage{subcaption}
\title{\LARGE \bf
AcousticDiffusion: Semantically Conditioned Audio-Guided \\ Diffusion Policy for Search-and-Rescue Assistance
}

\author{Iana Zhura$^{*}$, Didar Seyidov$^{*}$, Dmitrii Plotnikov, Hajira Amjad, \\ Miguel Altamirano Cabrera and Dzmitry Tsetserukou%
 \thanks{$^{*}$Equal contribution.}%
 \thanks{All authors are with the Skolkovo Institute of Science and Technology, Moscow, Russia. Email: {\tt\small \{ yana.zhura, didar.seyidov, dmitrii.plotnikov, hajira.amjad, m.altamirano, d.tsetserukou\}@skoltech.ru}}%
}

\begin{document}
\bstctlcite{IEEEexample:BSTcontrol}

\maketitle
\thispagestyle{empty}
\pagestyle{empty}

%%%%%%%%%%%%%%%%%%%%%%%%%%%%%%%%%%%%%%%%%%%%%%%%%%%%%%%%%%%%%%%%%%%%%%%%%%%%%%%%
\begin{abstract}
Navigating toward human callers is an important capability
for rescue robots operating where visual contact is
degraded or occluded. We present AcousticDiffusion,
a semantically conditioned, audio-guided diffusion policy
for human-directed navigation. A frozen pretrained audio
recognizer processes 10.24\,s windows, with speech gating
and distress-aware prioritization converting recognition
outputs into source-level navigation roles. Microphone-array
direction-of-arrival measurements are recursively integrated
into a robot-centric Bayesian bird's-eye-view belief field.
Ego-motion compensation aligns successive observations,
progressively constraining source position while preserving
bearing-induced range uncertainty. The semantic belief,
recent acoustic observations, audio features, and robot
state condition a diffusion model that generates waypoint
trajectories. On a synthetic-navigation validation set
using recorded audio, AcousticDiffusion achieves a mean
end-point bearing error of $11.20^\circ$, with $91.78\%$
of trajectories aligned within $30^\circ$ of the caller.
Distractor rejection ranges from $89.20\%$ to $98.99\%$,
and the policy favors a HELP-designated caller over a
competing speaker in $91.07\%$ of windows. Deployed online
on a ZSL-1 quadruped without additional
retraining, it achieves a mean bearing error of
$64.9^\circ$, compared with $98.2^\circ$ for A* and
$90.4^\circ$ for RRT, with a mean planner compute time
of $6.07$\,ms. Despite imperfect acoustic localization,
the reported mean final source distance is reduced from
$3.96$\,m for the classical planners using ODAS-derived (Open embedded Audition System)
guidance to $2.48$\,m, a $37.4\%$ improvement.
These results demonstrate the framework's ability
to translate uncertain acoustic observations into
closer approaches to human callers.
\end{abstract}

%%%%%%%%%%%%%%%%%%%%%%%%%%%%%%%%%%%%%%%%%%%%%%%%%%%%%%%%%%%%%%%%%%%%%%%%%%%%%%%%
%%%%%%%%%%%%%%%%%%%%%%%%%%%%%%%%%%%%%%%%%%%%%%%%%%%%%%%%%%%%%%%%%%%%%%%%%%%%%%%%
%%%%%%%%%%%%%%%%%%%%%%%%%%%%%%%%%%%%%%%%%%%%%%%%%%%%%%%%%%%%%%%%%%%%%%%%%%%%%%%%
 
% ──────────────────────────────────────────────────────────────
\section{INTRODUCTION}
 \begin{figure}[t]
    \centering
    \includegraphics[width=0.8\columnwidth]{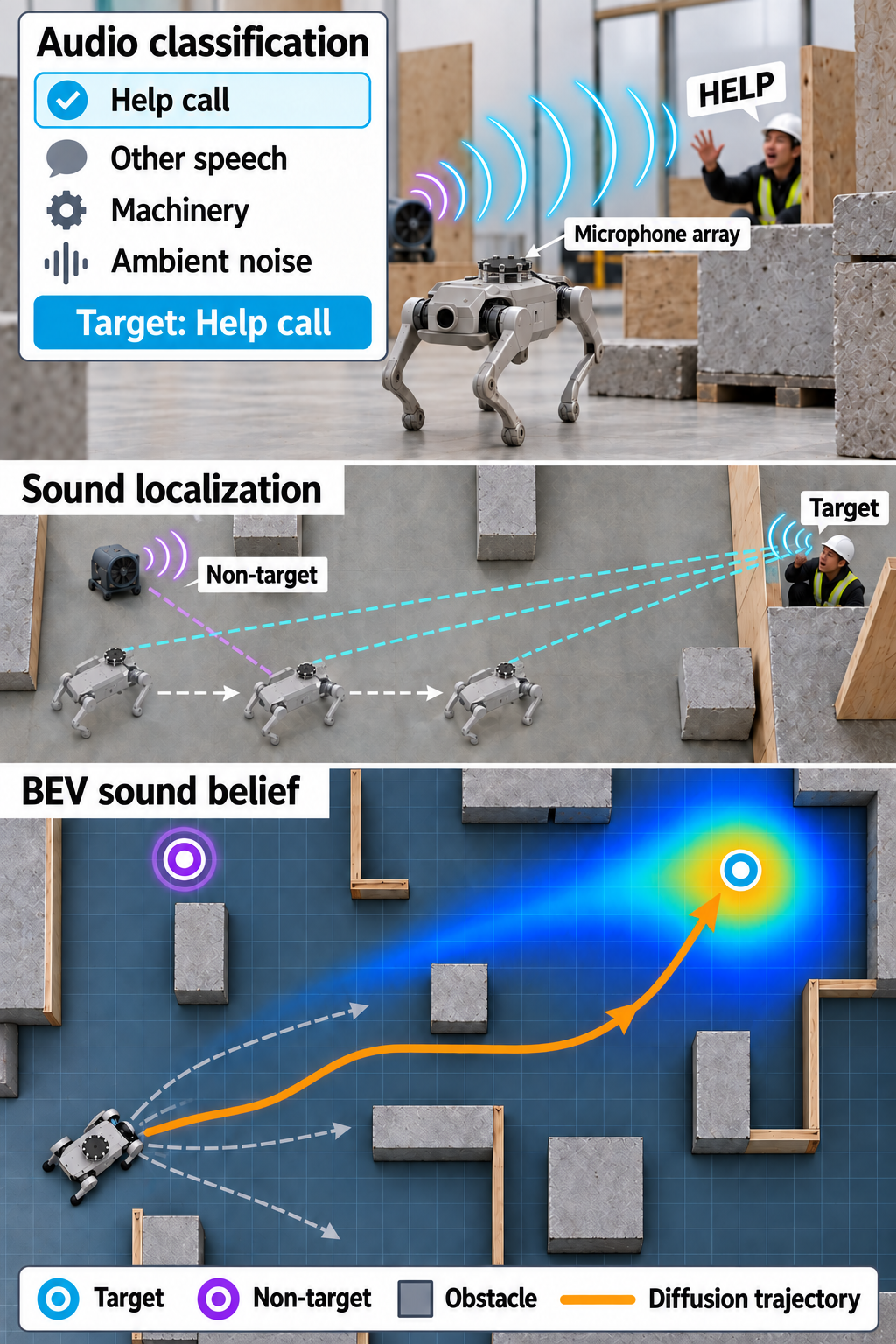}
    \caption{Overview of \textsc{AcousticDiffusion}. The quadruped robot
    identifies a task-relevant acoustic event among competing sound
    classes, accumulates direction-of-arrival observations into a
    bird's-eye-view sound belief, and generates a diffusion-based
    trajectory toward the target while avoiding obstacles.}
    \label{fig:echodiffusion_teaser}
\end{figure}

\label{sec:introduction}

Navigating toward a sound source is a natural and practical capability for
service and rescue robots operating in human environments. Spoken commands,
ringing phones, industrial alarms, and distress calls can reveal events and
people that are outside the robot's field of view or occluded by obstacles.
Unlike purely geometric acoustic navigation, however, rescue missions require
the robot to determine not only \emph{where} a sound originates, but also
\emph{which} sound should guide its motion. A person calling or screaming for
help may coexist with speech, machinery, alarms, reflections, and ambient
noise. The navigation policy must therefore recognize the semantic class of
each acoustic observation, select the task-relevant source, and generate a
safe trajectory toward it.

Sound-source navigation is further complicated by the bearing-only nature of
microphone-array observations. A single Direction-of-Arrival (DoA)
measurement constrains the source to a ray in free space rather than to a
point, providing no direct estimate of range. Recovering source position
requires observations from spatially separated robot poses or a strong
geometric prior. Robot motion is therefore not only a means of reaching the
source, but also an information-gathering action: changes in viewpoint create
acoustic parallax, allowing successive bearing measurements to triangulate the
source and reduce localization uncertainty.

Classical systems typically address this problem using an explicit acoustic
localization front end, such as GCC-PHAT~\cite{knapp1976gcc},
SRP-PHAT~\cite{dibiase2001srp}, or ODAS~\cite{grondin2019odas}, followed by a
separate navigation planner. Although this modular design is interpretable,
localization and control are treated as largely independent problems. The
planner generally acts on a point estimate after localization, rather than on
the evolving spatial uncertainty, and cannot directly use motion to resolve
bearing ambiguity. Moreover, geometric localization alone cannot determine
whether a detected source corresponds to a distress call, irrelevant speech,
machinery, or background noise.

% Recent learning-based methods have demonstrated the value of maintaining a closed acoustic sensorimotor loop~\cite{chen2020soundspaces,chen2021waypoints}. Nevertheless, many existing acoustic-navigation benchmarks use monaural or binaural audio rendered in simulation and assume that the navigation target is already known. Such assumptions do not directly address real microphone-array geometry, acoustic clutter, intermittent detections, competing sources, or semantic selection of a task-relevant sound. In parallel, diffusion models have emerged as expressive trajectory generators for contact-rich manipulation~\cite{chi2023diffusion} and mobile navigation~\cite{pearce2023imitating}. By representing a distribution over future waypoint sequences, diffusion policies can capture multiple feasible routes instead of committing to a single greedy action.

Recent learning-based approaches have demonstrated the value of coupling acoustic perception with embodied navigation \cite{gan2020look,chen2020soundspaces,chen2021waypoints,chen2021semantic}. These methods show that audio can provide navigation cues beyond the robot's visual field of view and that acoustic observations accumulated during motion can improve goal-directed navigation. However, most established benchmarks operate primarily in simulation using rendered monaural or binaural audio and formulate the task as reaching a designated sound-emitting goal. More recent work has investigated microphone-array-based navigation on physical robots \cite{rao2022listen} and sim-to-real transfer of audio-visual navigation policies \cite{chen2024sim2real}. Nevertheless, jointly selecting a task-relevant sound based on its semantic meaning, recursively resolving bearing-only localization uncertainty through robot motion, and directly conditioning trajectory generation on the resulting acoustic belief remain largely unexplored.

In parallel, diffusion models have emerged as expressive policies for sequential robot control, including contact-rich manipulation \cite{chi2023diffusion} and real-world mobile navigation \cite{sridhar2024nomad}. By representing a distribution over future actions or waypoints, diffusion policies are particularly well suited to navigation problems in which multiple feasible trajectories may satisfy the same perceptual objective.

We introduce \textbf{AcousticDiffusion}, a semantically conditioned, audio-guided diffusion policy for rescue navigation, including visually occluded environments. A frozen AudioSet-pretrained Audio Spectrogram Transformer (AST) processes 10.24\,s audio windows. Speech gating and distress-aware speaker prioritization convert its outputs into source-level navigation roles, retaining ordinary speech as a potential target. These semantics are integrated with microphone-array DoA measurements in a robot-centric Bayesian bird's-eye-view (BEV) field. Ego-motion compensation aligns evidence from successive viewpoints, allowing bearing-only observations to progressively constrain source position while preserving localization uncertainty. The semantic BEV representation, recent DoA and role tokens, audio features, and robot state jointly condition an $\epsilon$-prediction diffusion model that generates target-directed waypoint trajectories. The main contributions of this work are:

\begin{enumerate}
    \item A \textbf{semantically conditioned diffusion navigation framework} that adapts a frozen AudioSet-pretrained AST to source-level navigation through speech gating and distress-aware speaker prioritization. Rather than directly using predicted sound labels, this adaptation converts recognition scores into navigation roles while retaining ordinary speech as a potential target. The resulting semantic conditioning, Bayesian BEV belief, DoA tokens, and ego-state guide an $\epsilon$-prediction diffusion policy to generate target-directed waypoint trajectories.

    \item A \textbf{recursive semantic Bayesian BEV sound field} that accumulates bearing-only DoA evidence with ego-motion compensation and associates spatial evidence with source semantics. Successive observations progressively constrain source position, while spatial belief and uncertainty readouts condition trajectory generation. Conditioning ablations identify the BEV role maps as the principal semantic pathway supporting target alignment.

    \item A \textbf{ROS~2 implementation and real-robot evaluation} on a ZSL-1 quadruped, integrating microphone-array observations, semantic audio processing, odometry, diffusion trajectory generation, and LiDAR/IMU based collision avoidance. Experiments assess navigation toward task-relevant sound sources amid competing sounds and physical obstacles. The ROS~2 implementation and trained model checkpoints will be released upon acceptance.
\end{enumerate}
% ──────────────────────────────────────────────────────────────
% ============================================================
%  AcousticDiffusion — Section II: RELATED WORK (corrected, full-line format)
%  Drop this block into root.tex replacing the existing \section{RELATED WORK}
% ============================================================
 
\section{RELATED WORK}

\subsection{Acoustic Localization and Bearing-Only Tracking}

Classical microphone-array localization methods include
GCC-PHAT~\cite{knapp1976gcc}, which estimates inter-channel
time delays, and SRP-PHAT~\cite{dibiase2001srp}, which
evaluates steered responses over candidate source
locations. ODAS~\cite{grondin2019odas} integrates
localization, tracking, and separation for real-time
robot audition and supplies our DoA observations.

A single bearing does not determine source range;
localization requires additional geometric constraints,
such as observations from spatially separated robot
poses~\cite{nardone1984bof}. Particle
filters~\cite{arulampalam2002particle} provide one
approach to nonlinear, non-Gaussian tracking.
AcousticDiffusion instead accumulates bearing evidence
in a grid-based log-odds field, using odometry to align
successive observations in the robot frame.
This preserves multiple spatial hypotheses and
localization uncertainty for conditioning the
trajectory policy.

% \subsection{Sound-Source Navigation in Simulation}
\subsection{Audio-Guided and Audio-Visual Navigation}

% Chen et al.~\cite{chen2020soundspaces} introduced SoundSpaces, a large-scale benchmark for audio-visual navigation in which an agent must navigate to a sound-emitting goal in a realistic 3-D indoor environment. Their follow-up~\cite{chen2021waypoints} reformulated the task as waypoint prediction conditioned on an acoustic egomotion map, substantially improving sample efficiency and generalisation. Subsequent work has added semantic audio goals~\cite{majumder2021semantic}, multiple simultaneously active speakers~\cite{chen2022soundspaces2}, and zero-shot transfer of simulation-trained policies to real robot hardware~\cite{chen2024sim2real}.

Learning-based audio-guided navigation has received growing attention in embodied AI and robotics. Gan et al.~\cite{gan2020look} formulated audio-visual embodied navigation as the task of reaching a sound source from egocentric auditory and visual observations. SoundSpaces \cite{chen2020soundspaces} subsequently provided realistic acoustic simulation in scanned 3-D environments and established a widely used AudioGoal navigation benchmark. Chen et al.~\cite{chen2021waypoints} further introduced learned waypoint prediction together with a spatially grounded acoustic memory, allowing an agent to accumulate auditory information as it moves rather than acting only on the current observation. SoundSpaces~2.0~\cite{chen2022soundspaces2} extended this line of work with continuous, geometry-based acoustic simulation and configurable acoustic sensing.

% These benchmarks use monaural or binaurally rendered audio convolved with measured room impulse responses, which captures reverberation realistically but does not model a physical microphone array geometry, ODAS-style DoA parsing, or the clutter and phantom tracks that appear in real recordings. Notably, even the one paper that attempts sim-to-real transfer~\cite{chen2024sim2real} still relies on simulation-rendered binaural audio for training. AcousticDiffusion is designed for real array data from the start: it ingests ODAS SSL/SST streams directly, handles the near-vertical phantom tracks that appear in our reference session, and is validated against synthetic episodes whose DoA statistics are matched to those recordings, making it the first audio navigation system trained and evaluated entirely without simulation-rendered audio.

Beyond navigation toward continuously active sound sources, subsequent work has considered richer acoustic settings. Semantic Audio-Visual Navigation~\cite{chen2021semantic} introduced semantically grounded and temporally intermittent acoustic events, together with persistent multimodal memory that permits navigation after the sound has ceased. Move2Hear~\cite{majumder2021move2hear} demonstrated that agent motion can also be selected actively to improve separation of a target sound from competing sources. Sound Adversarial Audio-Visual Navigation \cite{yu2022soundadversarial} studied navigation in the presence of interfering acoustic sources, while Younes et al.~\cite{younes2023catch} considered moving targets, distractor sounds, and previously unheard acoustic events. These works demonstrate the importance of semantic reasoning, acoustic memory, and active perception when navigation takes place in complex sound scenes.

Most of these approaches are developed and evaluated primarily using simulated binaural observations. Work closer to physical robot audition includes Rao and Chowdhury~\cite{rao2022listen}, who demonstrated audio-aware navigation with a microphone-array-equipped mobile robot in indoor environments. Chen et al.~\cite{chen2024sim2real} subsequently studied sim-to-real transfer for audio-visual navigation through frequency-adaptive acoustic-field prediction and demonstrated navigation to sounding objects on a physical robot. These studies establish the feasibility of real-world audio-guided navigation, but do not jointly address semantic selection of a task-relevant source, recursive bearing-only localisation with explicit uncertainty, and generative trajectory prediction. AcousticDiffusion combines these elements by associating source semantics with tracked DoA observations, accumulating the selected bearing evidence into a robot-centric Bayesian belief, and conditioning trajectory generation directly on that evolving belief.

% \subsection{Diffusion Policies for Navigation}
\subsection{Diffusion Policies for Robot Navigation}

Diffusion Policy~\cite{chi2023diffusion} and diffusion-based
imitation learning~\cite{pearce2023imitating} generate
multimodal action sequences through conditional denoising.
For mobile robots, NoMaD~\cite{sridhar2024nomad} combines
visually guided navigation and goal-agnostic exploration.
When coupled with person detection or a suitable visual
goal, visually conditioned trajectory generation could
support approaching people for assistance; however,
direct visual localization requires visibility.
AcousticDiffusion instead uses acoustic semantics and
a recursive spatial belief derived from bearing-only
observations and ego-motion, providing guidance toward
human callers even when visual contact is unavailable.

\section{METHOD}
\label{sec:method}
\begin{figure*}[t]
    \centering
    \includegraphics[width=0.85\textwidth]{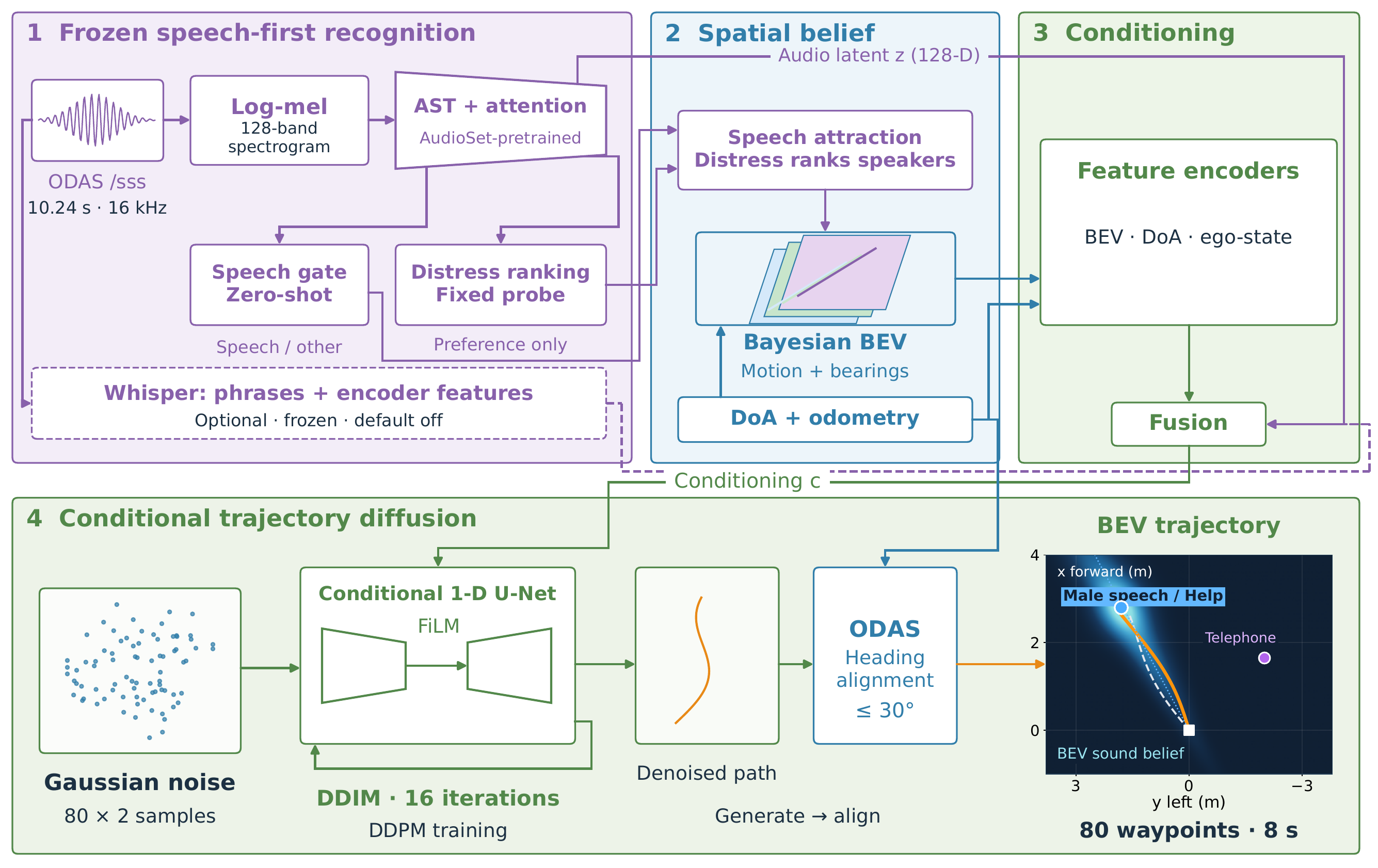}
    \caption{Overview of AcousticDiffusion. Pretrained audio recognition provides semantic predictions and audio features. Sound-source bearings and odometry update a Bayesian BEV belief. Fused     conditioning guides a 1-D U-Net that generates trajectories from Gaussian noise through 16 DDIM iterations. The BEV output and denoising stages are illustrative.}
    \label{fig:echodiffusion_architecture}
\end{figure*}
% Requires \usepackage{amsmath,amssymb} in the main document.

AcousticDiffusion generates trajectories toward human sound sources using frozen audio recognition, a recursive Bayesian bird's-eye-view (BEV) field, and a conditional diffusion policy (Fig.~\ref{fig:echodiffusion_architecture}). The system receives ODAS-separated audio, source-bearing observations, and robot odometry. Speech determines whether a source attracts the robot; distress provides a preference among speakers rather than a prerequisite for approach. The default configuration uses acoustic conditioning without a visual encoder. Recognizer parameters remain fixed while the trajectory policy is trained.

\subsection{Frozen Speech Recognition and Speaker Preference}
\label{sec:recognition}

\paragraph{Audio input and pretrained representation.}
Each tracked source is represented by a trailing 10.24\,s audio window sampled at 16\,kHz. Windows advance every 0.5\,s. The window duration and decision thresholds are loaded from the recognizer checkpoint, ensuring consistent offline and online preprocessing. Audio is downmixed, resampled, DC-corrected, adjusted to the prescribed duration, and RMS-normalized using the shared canonicalization pipeline. Shorter clips are padded; padding does not replace the information supplied by real audio. A 128-bin log-mel filterbank represents logarithmic sound energy over time across mel-frequency bands, using AST's 1024-frame input budget.

An AudioSet-pretrained Audio Spectrogram Transformer (AST) extracts acoustic features, with attention pooling providing a 128-dimensional latent $\mathbf{z}_{t,i}$ for source $i$ at time $t$. The backbone is frozen and no task-specific speech-recognizer training is performed in the current pipeline. Its pretrained AudioSet labels supply the speech score $s_{t,i}\in[0,1]$ directly, one scalar per tracked source per time step. The previous 13-class classifier and separate NONE/HAIL/HELP intent head are not constructed in the speech-first configuration.

\paragraph{Speech gating and distress preference.}
The system distinguishes other sounds, neutral speech, and distress calls. These outputs arise from two decisions with different roles: speech detection gates attraction, whereas distress ranks speech sources. The current checkpoint additionally contains an already-fitted distress probe, held fixed during policy training and inference. Speech gating is therefore zero-shot, while distress ranking uses fixed, previously fitted parameters.

For a detected speaker, distress modulates preference within a bounded interval $[\beta,1]$, where $\beta=0.6$ in the documented configuration. A neutral speaker therefore retains a nonzero attraction preference; missing distress lowers priority rather than converting the person into an obstacle. Sources are compared using speech evidence in logit space so that the weaker distress preference does not override the speech gate. Source roles are ATTRACT or NEUTRAL; no source receives an AVOID role. Calibrated thresholds are restored from the checkpoint rather than replaced by a generic threshold of 0.5.

% \paragraph{Optional verbal features.}
% A frozen Whisper-small branch can supply a help-phrase score and a mean-pooled encoder embedding. Phrase matching captures explicit verbal requests, while the embedding retains information beyond the recognized words. Both features can condition the policy. This branch is disabled for the default checkpoint and is enabled only when the policy's training episodes contain matching verbal features. No emotional interpretation is assumed solely from the presence of an encoder embedding.

\subsection{Recursive Bayesian Field and BEV Estimation}
\label{sec:bev}

\paragraph{Bearing evidence and temporal alignment.}
A single direction-of-arrival (DoA) observation constrains source direction but not range. For a BEV cell $\mathbf{q}=(x,y)$ and bearing $\theta_{t,i}$, the angular likelihood is represented by a von Mises ridge:
\begin{equation}
 \begin{aligned}
 \phi(\mathbf{q}) &= \operatorname{atan2}(y,x),\\
 \ell_{t,i}(\mathbf{q}) &\propto
 \exp\!\left\{\kappa_{t,i}
 [\cos(\phi(\mathbf{q})-\theta_{t,i})-1]\right\},
 \end{aligned}
\end{equation}
where $\phi(\mathbf{q})$ is the bearing of cell $\mathbf{q}$ as seen from the robot's own origin, i.e.\ the azimuth a source located at $\mathbf{q}$ would produce; $\theta_{t,i}$ is the measured DoA bearing for tracked source $i$ at time $t$; and $\ell_{t,i}(\mathbf{q})$ is the resulting angular likelihood assigned to cell $\mathbf{q}$, a ridge peaked where $\phi(\mathbf{q})=\theta_{t,i}$ and uniform in range. The concentration $\kappa_{t,i}$ depends on detection energy or activity (defined below); larger $\kappa_{t,i}$ sharpens the ridge around $\theta_{t,i}$, while $\kappa_{t,i}\to 0$ flattens it toward a uniform prior over bearing.

The concentration $\kappa_{t,i}$ depends on detection energy or activity. For a fixed direction, this observation model is uniform in range and does not manufacture a distance estimate from a single bearing.

The field accumulates evidence in robot-centric log-odds grids. Before incorporating current observations, the previous field is warped using the odometric pose increment and attenuated by temporal decay: 
\begin{equation}
 L_t^- = \rho_t\,\mathcal{W}_{\Delta\mathbf{p}_t}(L_{t-1}),
\end{equation}
where $L_{t-1}$ is the accumulated log-odds field from the previous time step and $L_t^-$ is its predicted value at time $t$, before that step's new bearing evidence is folded in. $\Delta\mathbf{p}_t$ is the odometric pose increment (translation and rotation) between $t-1$ and $t$, and $\mathcal{W}_{\Delta\mathbf{p}_t}(\cdot)$ is the rigid warp operator that re-expresses the previous field in the robot's new body frame under that motion, so that cells still refer to the same world locations after the robot has moved. $\rho_t\in(0,1]$ is a scalar temporal-decay (forgetting) factor applied at time $t$, implementing the field's evidence half-life: $\rho_t<1$ lets stale evidence fade so that a source which has moved does not leave a persistent ghost.

New angular evidence updates the aligned grids. Observations from spatially separated poses can then intersect, progressively constraining source range when robot motion provides sufficient parallax. Decay limits the persistence of obsolete evidence.

\paragraph{Semantic field construction.}
Untracked SSL potentials update the agnostic field without assigned semantic roles. Tracked SST bearings are associated with audio-derived roles through source identifiers and contribute to the semantic field. The speech-first configuration supplies attraction and neutral evidence without avoidance assignments. Inactive tracks and near-vertical detections are filtered while preserving the association between each bearing and its semantic information.

The online runner buffers ODAS callbacks and assembles observations at 10\,Hz, using at most the latest observation for each track in a tick. Odometry history is sampled on the same time base. Shared dataset and online window construction preserves BEV-channel layout, token ordering, pose conventions, and field-readout dimensions.

\paragraph{Spatial estimates.}
The policy receives accumulated BEV evidence and recent observation history, together with source-position and uncertainty readouts. A centroid estimate can be obtained by normalizing the log-odds field:
\begin{equation}
 \pi_t(\mathbf{q}) =
 \frac{\exp L_t(\mathbf{q})}
 {\sum_{\mathbf{q}'}\exp L_t(\mathbf{q}')},
 \qquad
 \hat{\mathbf{s}}_t = \sum_{\mathbf{q}}\pi_t(\mathbf{q})\mathbf{q}.
\end{equation}
$L_t(\mathbf{q})$ is the fully updated log-odds field at cell $\mathbf{q}$ and time $t$ (i.e.\ $L_t^-$ after that step's bearing evidence has been incorporated). $\pi_t(\mathbf{q})$ is the corresponding posterior probability mass at cell $\mathbf{q}$, obtained by a softmax over the log-odds field, so that $\pi_t$ is a proper probability distribution over the grid ($\sum_{\mathbf{q}}\pi_t(\mathbf{q})=1$); $\mathbf{q}'$ is a dummy variable ranging over all grid cells, used only for that normalizing sum. $\hat{\mathbf{s}}_t$ is the resulting centroid estimate of the source position at time $t$: the probability-weighted mean grid coordinate.

Normalizing log-odds retains spatial contrast that may be lost after sigmoid saturation. Spread and confidence accompany the spatial readout. The field estimate remains distinct from the learned source estimate produced by the policy's auxiliary head.

\subsection{Conditional Trajectory Generation}
\label{sec:diffusion}

\paragraph{Policy conditioning.}
A CNN encodes the BEV belief with spatial pooling. A DoA encoder embeds source tokens and role information using masked mean and maximum aggregation followed by temporal processing. An ego-state encoder processes pose history and field readouts. These features and the pooled audio latent form the conditioning vector:
\begin{equation}
 \begin{aligned}
 \mathbf{c}_t = f_{\mathrm{fuse}}\!\big(&
 f_{\mathrm{BEV}}(B_t), f_{\mathrm{DoA}}(D_t),\\
 &f_{\mathrm{ego}}(E_t),\mathbf{z}_t\big).
 \end{aligned}
\end{equation}
$B_t$ is the BEV belief input at time $t$ (the accumulated field $L_t$ together with recently re-rendered instantaneous ray maps) and $f_{\mathrm{BEV}}(\cdot)$ is the CNN encoder above that maps it to a pooled feature grid. $D_t$ is the set of per-source DoA tokens at time $t$ (bearing plus role information for every currently tracked source) and $f_{\mathrm{DoA}}(\cdot)$ is the DoA encoder that aggregates them via masked mean/max pooling and temporal processing. $E_t$ is the ego-state input at time $t$ (recent pose history and the field's spread/confidence readouts) and $f_{\mathrm{ego}}(\cdot)$ is its encoder. $\mathbf{z}_t$ is the pooled audio latent carried forward from the associated source's $\mathbf{z}_{t,i}$ (Sec.~\ref{sec:recognition}). $f_{\mathrm{fuse}}(\cdot)$ is the fusion layer that combines all four encoded streams into the single conditioning vector $\mathbf{c}_t$ used below.

For a verbally conditioned checkpoint, the Whisper phrase score and encoder embedding are additional inputs. The default policy does not receive newly enabled verbal features at deployment.

\paragraph{Diffusion training and inference.}
The policy represents a normalized trajectory as $\boldsymbol{\tau}_0\in\mathbb{R}^{80\times2}$. Training perturbs an expert trajectory with Gaussian noise at diffusion timestep $k$:
\begin{equation}
 \boldsymbol{\tau}_k =
 \sqrt{\bar\alpha_k}\boldsymbol{\tau}_0 +
 \sqrt{1-\bar\alpha_k}\boldsymbol{\epsilon},
 \qquad \boldsymbol{\epsilon}\sim\mathcal{N}(\mathbf{0},\mathbf{I}).
\end{equation}
where $\boldsymbol{\tau}_0$ is the clean expert trajectory (80 waypoints in the plane) and $k$ indexes the diffusion timestep, i.e.\ how much noise has been mixed in. $\bar\alpha_k\in(0,1]$ is the standard DDPM
cumulative noise-schedule coefficient at step $k$ (the product of per-step signal-retention factors), so that $\bar\alpha_k$ decreases from near 1 at $k{=}0$ (clean trajectory) toward 0 as $k$ grows (pure noise). $\boldsymbol{\epsilon}$ is the Gaussian noise sample actually added, drawn once per training example from a standard normal $\mathcal{N}(\mathbf{0},\mathbf{I})$, and $\boldsymbol{\tau}_k$ is the resulting noised trajectory at step $k$ that the network is trained to denoise.

A conditional 1-D U-Net uses feature-wise linear modulation (FiLM) and skip connections to predict the injected noise. The noise-prediction objective is:
\begin{equation}
 \mathcal{L}_{\mathrm{diff}} =
 \mathbb{E}\!\left[
 \left\|\boldsymbol{\epsilon}-
 \epsilon_\theta(\boldsymbol{\tau}_k,k,\mathbf{c}_t)\right\|_2^2
 \right].
\end{equation}
where $\epsilon_\theta(\boldsymbol{\tau}_k,k,\mathbf{c}_t)$ is the U-Net's predicted noise, a function of the noised trajectory $\boldsymbol{\tau}_k$, the timestep $k$, and the conditioning vector $\mathbf{c}_t$, with $\theta$ denoting the network's trainable weights (FiLM-modulated by $k$ and $\mathbf{c}_t$). $\mathbb{E}[\cdot]$ is the expectation over training trajectories, sampled timesteps $k$, and noise draws $\boldsymbol{\epsilon}$, and $\|\cdot\|_2^2$ is the squared Euclidean norm over the trajectory's 80 waypoints. $\mathcal{L}_{\mathrm{diff}}$ is the resulting scalar training loss minimized over $\theta$.

An auxiliary source-regression head provides a learned source-position estimate when source supervision is available. Navigation training uses synthetic expert trajectories; frozen recognition supplies conditioning without being updated by the diffusion loss.

At inference, generation starts from Gaussian noise and uses 16 DDIM iterations. The resulting 80 waypoints span 8\,s at 10\,Hz. 
% The ROS~2 node
% replans every 0.5\,s, restores normalization and window geometry from the
% policy checkpoint, and publishes the path in the robot frame. It also
% publishes the learned speaker position and field-based estimates.

\paragraph{Post-sampling heading alignment.}
The default online configuration refines trajectory direction using ODAS after denoising. Among eligible tracks with attraction posterior at least 0.05, it selects the bearing nearest the generated path heading within a $30^\circ$ gate. If a track is selected, the entire trajectory is rotated about the robot by the heading difference:
\begin{equation}
 \tilde{\mathbf{q}}_j = R(\Delta\theta)\mathbf{q}_j,
 \qquad j=1,\ldots,80.
\end{equation}
$\mathbf{q}_j$ is the $j$-th of the 80 waypoints produced by DDIM sampling, in the robot frame. $\Delta\theta$ is the heading-correction angle: the difference between the selected ODAS track's bearing and the generated path's own heading, as selected above. $R(\Delta\theta)$ is the $2\times2$ rotation matrix by angle $\Delta\theta$ about the robot's origin, and $\tilde{\mathbf{q}}_j$ is the corresponding rotated waypoint; applying the same rotation to every $j$ reorients the whole trajectory toward the selected source without changing its shape or speed profile.

% Otherwise, the trajectory is unchanged. This operation preserves path
% shape and length, but does not itself guarantee collision avoidance.

% Depth-sensor measurements provide complementary obstacle information
% for the robot's collision-avoidance component; depth is not represented
% as a visual-encoder input to this acoustic policy. The depth-processing
% and path-validation implementation is separate from the architecture
% described here.
\section{EXPERIMENTS AND RESULTS}
\subsection{Dataset Construction}
\label{sec:dataset}

Speech recordings are drawn from LibriSpeech~\cite{panayotov2015librispeech}, while emotional speech is obtained from RAVDESS~\cite{livingstone2018ravdess}. Fearful speech serves as a proxy for distress; these acted utterances are not literal requests for assistance. Audio is processed in 10.24\,s windows, with shorter recordings padded. Successive clips are concatenated without overlapping crossfades.

Synthetic scenes combine a designated caller, competing speakers, and non-speech distractors. A potential-field expert provides trajectory supervision. Simulated DoA observations are modeled on the detection patterns observed in real ODAS recordings collected using a microphone array mounted on our ZSL-1 quadruped.
The simulation reproduces clutter detections, intermittent source tracks, and near-vertical phantom tracks. The controlled speech-first configuration uses negligible inter-source leakage. Thus, real recordings inform the acoustic observation model, while navigation trajectories and their supervision remain synthetic.

The evaluated validation partition contains 414 windows: 296 with a HELP-designated caller and 118 without a caller. Caller windows comprise 160 female-speech and 136 male-speech examples; 112 additionally contain a competing human speaker. These labels describe the evaluation sources, rather than outputs of a trained multiclass recognition head. All navigation-policy ablations use this validation partition.

\subsection{Human-Directed Navigation}
\label{sec:human_directed}

We evaluate whether AcousticDiffusion follows callers while ignoring distracting sounds. Caller alignment measures end-point directions within $45^\circ$ of the caller. Distractor rejection counts source--window pairs where the distractor is not both nearest in bearing to the end-point and within $20^\circ$ of it. These metrics assess directional selection, not recognition accuracy or physical arrival.
% Preamble

Table~\ref{tab:class_selection} shows caller alignment above $96\%$ and distractor rejection of $89.20$--$98.99\%$. With a competing speaker, the policy favors the HELP-designated caller in $91.07\pm0.89\%$ of windows, supporting learned human-directed behavior. Without a caller, mean seed-wise median end-point displacement falls from $2.86$ to $0.27$\,m, near the expert's $0.24$\,m; however, only $1.69\%$ of predictions remain within $0.05$\,m, indicating reduced motion rather than reliable abstention. Improved emotion recognition could further strengthen caller prioritization; the current system uses distress as a preference, not a prerequisite for approaching a person.
\begin{figure*}[t]
    \centering
    \includegraphics[width=0.85\textwidth]{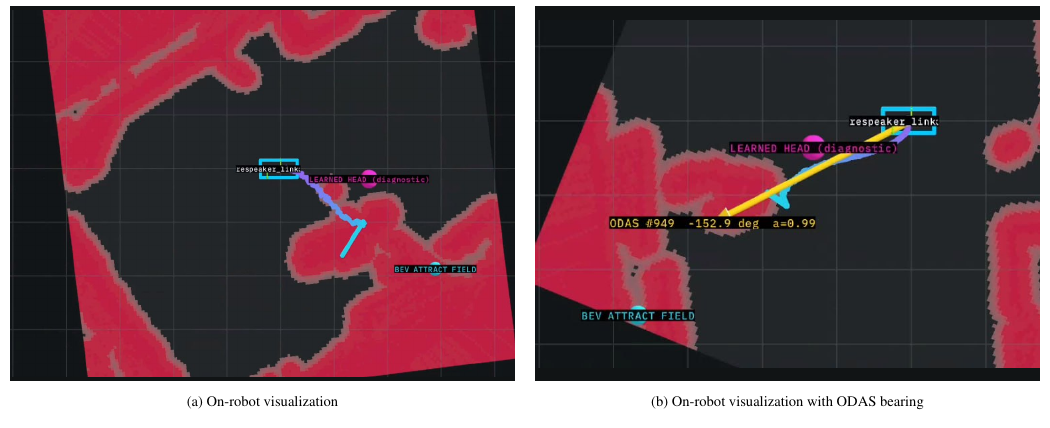}
    \caption{Qualitative visualizations from the
    AcousticDiffusion deployment on the quadruped robot.
    Both panels show the microphone-array frame
    (\texttt{respeaker\_link}), the displayed trajectory,
    the learned source-head diagnostic (magenta), and
    the BEV attraction-field estimate (red), overlaid
    on the local map. Panel (b) additionally shows
    the ODAS bearing observation in yellow.}
    \label{fig:robot_qualitative}
\end{figure*}

\subsection{Qualitative Real-Robot Results}
\label{sec:robot_qualitative}

Fig.~\ref{fig:robot_qualitative} presents on-robot visualizations of AcousticDiffusion. The displayed trajectories and acoustic estimates illustrate the integration of microphone-array observations, Bayesian spatial belief, and learned trajectory generation within the ROS~2 system. The learned source-head and BEV estimates can differ, highlighting the distinction between the two localization outputs. These examples demonstrate on-robot operation and successful sound source classification and localization. 
\begin{table}[t]
    \centering
    \caption{Human-directed navigation.
    Rates are mean $\pm$ sample standard deviation
    over three sampling seeds. $N$ counts windows
    containing each class; classes may co-occur.
    Laughter is treated as a non-speech distractor.}
    \label{tab:class_selection}
    \small
    \setlength{\tabcolsep}{5pt}
    \begin{tabular}{lrc}
        \toprule
        Source class & $N$ & Rate (\%) $\uparrow$ \\
        \midrule
        \multicolumn{3}{l}{Caller alignment within $45^\circ$} \\
        Female speech & 160 & $97.29\pm1.30$ \\
        Male speech   & 136 & $96.32\pm0.74$ \\
        \midrule
        \multicolumn{3}{l}{Distractor rejection} \\
        Domestic sounds    & 71  & $89.20\pm0.81$ \\
        Laughter           & 72  & $98.61\pm0.00$ \\
        Music              & 99  & $95.96\pm0.00$ \\
        Musical instrument & 66  & $98.99\pm0.87$ \\
        Telephone          & 108 & $95.99\pm1.07$ \\
        \bottomrule
    \end{tabular}
\end{table}

\subsection{Ablation: Diffusion Prediction Target}
\label{sec:ablation_prediction}

We compare noise prediction ($\epsilon$) and direct trajectory prediction ($\mathbf{x}_0$) with identical data, architecture, conditioning, and 100-epoch training. The recognizer remains frozen; evaluation uses final EMA checkpoints and 16-step DDIM without ODAS steering. 

Bearing error measures the angular difference between end-point and caller directions. Caller-directed rate identifies the caller as the nearest source in bearing; within-$30^\circ$ measures alignment within that tolerance. Neither implies physical arrival. ADE and FDE measure mean waypoint and final-point Euclidean errors against the expert, respectively. Relative roughness divides predicted roughness by expert roughness, with one indicating equal roughness.

\begin{table}[t]
    \centering
    \caption{Prediction-target ablation. Alignment values
    are mean $\pm$ sample standard deviation over three
    sampling seeds, not training runs. ADE, FDE, and
    roughness use 256 validation windows.}
    \label{tab:prediction_ablation}
    \small
    \setlength{\tabcolsep}{4pt}
    \begin{tabular}{lcc}
        \toprule
        Metric & $\mathbf{x}_0$ & $\epsilon$ \\
        \midrule
        Bearing error ($^\circ$) $\downarrow$
            & $17.42\pm0.07$ & $\mathbf{11.20}\pm0.33$ \\
        Caller-directed (\%) $\uparrow$
            & $87.05\pm0.39$ & $\mathbf{89.64}\pm0.98$ \\
        Within $30^\circ$ (\%) $\uparrow$
            & $80.86\pm0.20$ & $\mathbf{91.78}\pm1.67$ \\
        \midrule
        ADE (m) $\downarrow$
            & $\mathbf{0.285}$ & $0.329$ \\
        FDE (m) $\downarrow$
            & $0.846$ & $\mathbf{0.725}$ \\
        Roughness / expert $\downarrow$
            & $136.29$ & $\mathbf{76.98}$ \\
        \bottomrule
    \end{tabular}
\end{table}

Table~\ref{tab:prediction_ablation} reveals a trade-off between imitation accuracy and acoustic alignment: $\mathbf{x}_0$ prediction achieves lower ADE, whereas $\epsilon$ prediction reduces bearing error by $35.7\%$, improves within-$30^\circ$ alignment by $10.92$ percentage points, and lowers FDE. It also reduces roughness by $43.5\%$, although both variants remain substantially rougher than the expert. We therefore adopt $\epsilon$ prediction for its better target alignment and end-point accuracy; dependence on individual conditioning inputs is examined separately. 

\subsection{Ablation: Conditioning Dependence}
\label{sec:ablation_conditioning}

We remove conditioning inputs at inference time from a fixed $\epsilon$-prediction checkpoint, replacing them with unobserved priors or zeros. Speech removal affects token roles, the recognizer latent, BEV role maps, and the field role readout; ODAS removal affects DoA tokens, all BEV maps, and the field estimate. This tests input dependence without retraining.

\begin{table}[t]
    \centering
    \caption{Conditioning ablation: mean $\pm$ sample
    standard deviation over three sampling seeds.
    Bearing denotes end-point angular error to the caller.}
    \label{tab:conditioning_ablation}
    \small
    \setlength{\tabcolsep}{3pt}
    \begin{tabular}{lcc}
        \toprule
        Conditioning & Bearing ($^\circ$) $\downarrow$
        & Within $30^\circ$ (\%) $\uparrow$ \\
        \midrule
        Full
            & $11.20\pm0.33$ & $91.78\pm1.67$ \\
        No speech (prior)
            & $20.03\pm0.43$ & $79.84\pm0.70$ \\
        No speech (zero)
            & $24.16\pm0.76$ & $73.25\pm1.66$ \\
        No ODAS (prior)
            & $46.76\pm1.30$ & $32.58\pm0.97$ \\
        No ODAS (zero)
            & $24.16\pm0.85$ & $79.69\pm2.09$ \\
        Past motion only
            & $48.27\pm1.04$ & $25.14\pm2.16$ \\
        \midrule
        \multicolumn{3}{l}{Individual component removal} \\
        Token roles
            & $11.27\pm0.28$ & $90.77\pm1.67$ \\
        Recognizer latent
            & $10.55\pm0.04$ & $92.23\pm0.68$ \\
        BEV role maps
            & $19.44\pm0.45$ & $81.06\pm1.21$ \\
        Field role readout
            & $11.65\pm0.40$ & $91.33\pm1.86$ \\
        \bottomrule
    \end{tabular}
\end{table}

Removing speech or ODAS conditioning degrades alignment under both replacement schemes, while past motion alone reduces within-$30^\circ$ alignment from $91.78\%$ to $25.14\%$. Among individual semantic components, BEV role-map removal causes the largest degradation, identifying spatial semantics as the principal tested semantic pathway. Removing the recognizer latent slightly improves alignment but increases relative roughness from $77.59$ to $102.23$. We retain full conditioning for quadruped experiments.

\section{REAL-ROBOT EVALUATION}
\subsection{Robot Setup}
Experiments used a ZSL-1 quadruped equipped with an NVIDIA Jetson Orin NX, a Livox MID-360 LiDAR/IMU, and a 16-kHz Seeed reSpeaker XVF3800 four-microphone array, with ODAS~2.1 supplying direction-of-arrival tracks and separated audio. The array centre sits $0.25$~m behind and $0.30$~m above \texttt{base\_link}, rotated $180^\circ$ in yaw. All perception, inference, and navigation ran on-board, with no external compute.

Localization fused motor odometry with LiDAR--IMU measurements via planar NDT-OMP scan matching against a prior point-cloud map. Navigation used LiDAR costmaps with Nav2's MPPI controller, capped at $0.35$~m/s linear and $0.50$~rad/s angular speed.

An external Vicon system, tracking the microphone array and multiple acoustic emitters, supplied ground truth for evaluation only. Vicon and map coordinates were aligned per run; since emitter markers had zero horizontal offset from the physical sources, only height offsets were applied.

\subsection{Source Separation Configuration}

Microphone geometry was verified against manufacturer coordinates, with approximately 6.6\,cm inter-microphone spacing. We evaluated delay-and-sum (DS) beamforming and geometric source separation with decorrelation (DGSS), each with and without ODAS spectral-masking post-filtering to suppress residual noise and inter-source interference. ODAS's Kalman filter tracked source directions over time, separately from the audio post-filter.

Each configuration was evaluated with two active speakers using mean pairwise correlation between active output channels as a proxy for cross-talk. DS and DGSS required separate recording sessions and produced two and three active output channels, respectively, due to dynamic slot assignment, limiting direct comparison.

Post-filtering reduced correlation from $0.393$ to $0.001$ for DS and from $0.376$ to $0.061$ for DGSS. Without post-filtering, the methods produced similar correlations. We adopt DGSS with post-filtering, which reduced correlation by approximately $84\%$ relative to DGSS alone, although DS with post-filtering achieved the lowest measured correlation.

\subsection{ODAS-Coupled Baselines}
\label{sec:baseline_implementation}

A* and RRT* use the same active ODAS SST track identifier logged by AcousticDiffusion, avoiding independent source selection. At each replanning cycle, the robot origin serves as the start, and the track's reported robot-frame Cartesian coordinates define the goal. A* searches an 8-connected grid at $0.10$\,m resolution with a Euclidean heuristic; RRT* uses nearest-neighbor extension and cost-based rewiring within a bounded local region. Replanning is requested at approximately $1.5$\,Hz, independently of the recognizer's 10.24\,s analysis window.

\subsection{Comparison with Classical Planners}
\label{sec:baseline_comparison}
\begin{table}[t]
    \centering
    \caption{Comparison on trials S1--S3.
    AD denotes AcousticDiffusion. Windowed error is
    the mean of the minimum bearing errors within
    10.24\,s windows. Compute time measures planner
    execution, not end-to-end sensing latency.
    Bold indicates the best value, excluding the
    descriptive replanning period.}
    \label{tab:baseline_comparison}
    \small
    \setlength{\tabcolsep}{3pt}
    \begin{tabular}{lrrr}
        \toprule
        Metric & AD (ours) & A* & RRT \\
        \midrule
        \multicolumn{4}{l}{\textit{Bearing alignment}} \\
        Mean error ($^\circ$) $\downarrow$
            & \textbf{64.9} & 98.2 & 90.4 \\
        Windowed error ($^\circ$) $\downarrow$
            & \textbf{47.2} & 72.8 & 59.8 \\
        Windows $\leq15^\circ$ (\%) $\uparrow$
            & \textbf{32} & 18 & 25 \\
        \midrule
        \multicolumn{4}{l}{\textit{Planner compute and timing}} \\
        Mean compute (ms) $\downarrow$
            & 6.07 & \textbf{0.15} & 23.6 \\
        P95 compute (ms) $\downarrow$
            & 12.67 & \textbf{0.23} & 25.2 \\
        Median replan period (ms)
            & 645.7 & 682.1 & 682.1 \\
        \midrule
        \multicolumn{4}{l}{\textit{Final source distance}} \\
        Mean (m) $\downarrow$
            & \textbf{2.48} & 3.96 & 3.96 \\
        Median (m) $\downarrow$
            & \textbf{2.48} & 3.42 & 3.42 \\
        \bottomrule
    \end{tabular}
\end{table}
Table~\ref{tab:baseline_comparison} compares  AcousticDiffusion with A* and RRT on trials S1--S3, using the same acoustic bearing estimates and costmaps. We report instantaneous bearing error and the minimum bearing error within each 10.24\,s window, averaged over windows. The window duration matches the audio recognizer's analysis interval, allowing assessment over the same temporal horizon used for semantic conditioning. The windowed metric measures whether accurate alignment occurs during that interval, rather than whether it is maintained throughout. We additionally report the fraction of windows containing an alignment error of at most $15^\circ$.

AcousticDiffusion achieves the lowest instantaneous bearing error ($64.9^\circ$) and windowed error ($47.2^\circ$), compared with $98.2^\circ$ and $72.8^\circ$ for A*, and $90.4^\circ$ and $59.8^\circ$ for RRT. It reaches $15^\circ$ alignment in $32\%$ of windows, versus $18\%$ and $25\%$, respectively. These results indicate improved directional guidance, although the remaining errors show that accurate alignment is not consistently maintained.  

AcousticDiffusion's mean planner compute time is $6.07$\,ms, faster than RRT's $23.6$\,ms but slower than A*'s $0.15$\,ms on this costmap. Its P95 compute time of $12.67$\,ms remains well below the observed median replanning period of roughly $650$\,ms, as do the compute times of both baselines, so none of the three planners is compute bound at the replanning rate evaluated here. Finally, the reported mean final source distance decreases from $3.96$ to $2.48$\,m, a $37.4\%$ reduction. Together, these measurements favor AcousticDiffusion in directional alignment and final proximity within this evaluation.

\section{CONCLUSION AND FUTURE WORK}

We presented AcousticDiffusion, a speech-guided navigation framework combining pretrained audio recognition, a semantic Bayesian BEV sound field, and diffusion-based trajectory generation. On the validation set, the policy achieves a mean end-point bearing error of $11.20^\circ$, with $91.78\%$ of trajectories aligned within $30^\circ$ of the caller. Distractor rejection ranges from $89.20\%$ to $98.99\%$, and the HELP-designated caller is favored over a competing speaker in $91.07\%$ of windows. Deployed online on the quadruped without additional retraining, the policy achieves a mean bearing error of $64.9^\circ$, compared with $98.2^\circ$ for A* and $90.4^\circ$ for RRT, with a mean planner compute time of $6.07$\,ms. The larger real-robot error highlights the remaining simulation-to-reality gap.

Future work will fuse acoustic and visual features to improve localization and obstacle-aware navigation: audio guides motion toward occluded callers, while vision supplies spatial context. Conditional diffusion can integrate these complementary cues and represent multiple plausible trajectories under uncertainty. We will investigate whether this fusion can reduce reliance on dedicated depth sensing and explicit maps by learning collision-aware trajectories directly from observations.

\bibliographystyle{IEEEtran}
\bibliography{IEEEexample}

@IEEEtranBSTCTL{IEEEexample:BSTcontrol,
  CTLuse_forced_etal       = "yes",
  CTLmax_names_forced_etal = "6",
  CTLnames_show_etal       = "1"
}

@inproceedings{knapp1976gcc,
   author    = {Knapp, C. and Carter, G.},
   title     = {The Generalized Correlation Method for Estimation of Time Delay},
   booktitle = {IEEE Transactions on Acoustics, Speech, and Signal Processing},
   year      = {1976},
   volume    = {24},
   number    = {4},
   pages     = {320--327},
   doi       = {10.1109/TASSP.1976.1162830}
 }

@inproceedings{dibiase2001srp,
   author    = {DiBiase, J.H. and Silverman, H.F. and Brandstein, M.S.},
   title     = {Robust Localization in Reverberant Rooms},
   bookTitle = {Microphone Arrays: Signal Processing Techniques and Applications},
   year      = {2001},
   pages     = {157--180},
   doi       = {10.1007/978-3-662-04619-7_8}
 }

@inproceedings{panayotov2015librispeech,
  title={Librispeech: an asr corpus based on public domain audio books},
  author={Panayotov, Vassil and Chen, Guoguo and Povey, Daniel and Khudanpur, Sanjeev},
  booktitle={2015 IEEE international conference on acoustics, speech and signal processing (ICASSP)},
  pages={5206--5210},
  year={2015},
  organization={IEEE}
}

@article{livingstone2018ravdess,
  title={The Ryerson Audio-Visual Database of Emotional Speech and Song (RAVDESS): A dynamic, multimodal set of facial and vocal expressions in North American English},
  author={Livingstone, Steven R and Russo, Frank A},
  journal={PloS one},
  volume={13},
  number={5},
  pages={e0196391},
  year={2018},
  publisher={Public Library of Science San Francisco, CA USA}
}

@article{grondin2019odas,
   author    = {Grondin, Fran{\c{c}}ois and L{\'e}tourneau, Dominic and
             Godin, C{\'e}dric and Lauzon, Jean-Samuel and
             Vincent, Jonathan and Michaud, Simon and Faucher, Samuel and
             Michaud, Fran{\c{c}}ois},
   title     = {{ODAS}: Open embeddeD Audition System},
   journal   = {Frontiers in Robotics and AI},
   volume    = {9},
   pages     = {1--9},
   year      = {2022},
   doi       = {10.3389/frobt.2022.854444}
}

@inproceedings{chen2020soundspaces,
   author    = {Chen, Changan and Jain, Unnat and Schissler, Carl and Gari, Sebastia Vicenc Amengual and Al-Halah, Ziad and Ithapu, Vamsi Krishna and Robinson, Philip and Grauman, Kristen},
   title     = {{SoundSpaces}: Audio-Visual Navigation in 3{D} Environments},
   pages     = {17–36},
   doi       = {10.1007/978-3-030-58539-6_2},
   booktitle = {Computer Vision – ECCV 2020: European Conf.},
   year      = {2020}
 }

@inproceedings{chen2021semantic,
   author     = {Chen, C. and Al-Halah, Z. and Grauman, K.},
   title      = {Semantic Audio-Visual Navigation},
   booktitle  = {Proc. IEEE/CVF Conf. on Computer Vision and Pattern Recognition (CVPR)},
   year       = {2021},
   pages      = {15516--15525},
   doi        = {10.1109/CVPR46437.2021.01526}
 }

@inproceedings{majumder2021move2hear,
   author     = {Majumder, S. and Al-Halah, Z. and Grauman, K.},
   title      = {Move2Hear: Active Audio-Visual Source Separation},
   booktitle  = {Proc. IEEE/CVF Int. Conf. on Computer Vision (ICCV)}, 
   year       = {2021},
   pages      = {275-285},
   doi        = {10.1109/ICCV48922.2021.00034}
 }

@inproceedings{chen2022soundspaces2,
   author     = {Chen, Changan and Schissler, Carl and Garg, Sanchit and Kobernik, Philip and Clegg, Alexander and Calamia, Paul and Batra, Dhruv and Robinson, Philip and Grauman, Kristen},
   title      = {{SoundSpaces} 2.0: A Simulation Platform for Visual-Acoustic Learning},
   booktitle  = {Proc. Int. Conf. on Neural Information Processing Systems (NeurIPS)},
   pages      = {1--16},
   year       = {2022}
 }

@article{chi2023diffusion,
   author     = {Chi, Cheng and Xu, Zhenjia and Feng, Siyuan and Cousineau, Eric and Du, Yilun and Burchfiel, Benjamin and Tedrake, Russ and Song, Shuran},
   editor     = {Kostas Bekris, Kris Hauser, Sylvia Herbert and Jingjin Yu},
   title      = {Diffusion policy: Visuomotor policy learning via action diffusion},
   year       = {2025},
   volume     = {44},
   number     = {10–11},
   doi        = {10.1177/02783649241273668},
   journal    = {Int. Journal of Robotics Research},
   month      = sep,
   pages      = {1684–1704}
}

@inproceedings{pearce2023imitating,
   author     = {Pearce, T. and Rashid, T. and Ciosek, K. and Lengyel, M.},
   title      = {Imitating Human Behaviour with Diffusion Models},
   booktitle  = {Proc. Int. Conf. on Learning Representations (ICLR)},
   year       = {2023},
   pages      = {1--24}
 }

@article{nardone1984bof,
   author    = {Nardone, S.C. and Lindgren, A.G. and Gong, K.F.},
   title     = {Fundamental Properties and Performance of Conventional
                Bearings-Only Target Motion Analysis},
   journal   = {IEEE Transactions on Automatic Control},
   year      = {1984},
   volume    = {29},
   number    = {9},
   pages     = {775-787},
   doi       = {10.1109/TAC.1984.1103664}
 }

@inproceedings{arulampalam2002particle,
   author    = {Arulampalam, M.S. and Maskell, S. and Gordon, N. and Clapp, T.},
   title     = {A Tutorial on Particle Filters for Online Nonlinear/Non-Gaussian
                Bayesian Tracking},
   booktitle = {IEEE Transactions on Signal Processing},
   year      = {2002},
   volume    = {50},
   number    = {2},
   pages     = {174-188},
   doi       = {10.1109/78.978374}
 }

@inproceedings{gan2020look,
   author    = {Gan, Chuang and Zhang, Yiwei and Wu, Jiajun and Gong, Boqing and Tenenbaum, Joshua B.},
   title     = {Look, Listen, and Act: Towards Audio-Visual Embodied Navigation},
   booktitle = {Proc. IEEE Int. Conf. on Robotics and Automation (ICRA)},
   year      = {2020},
   pages     = {9701-9707},
   doi       = {10.1109/ICRA40945.2020.9197008}
 }

@inproceedings{chen2024sim2real,
  author    = {Chen, Changan and Ramos, Jordi and Tomar, Anshul and Grauman, Kristen},
  title     = {Sim2Real Transfer for Audio-Visual Navigation with Frequency-Adaptive Acoustic Field Prediction},
  booktitle = {IEEE/RSJ International Conference on Intelligent Robots and Systems (IROS)},
  pages     = {8595--8602},
  year      = {2024}
}

@inproceedings{rao2022listen,
  author    = {Rao, Pratyaksh P. and Chowdhury, Abhra Roy},
  title     = {Learning to Listen and Move: An Implementation of Audio-Aware
               Mobile Robot Navigation in Complex Indoor Environment},
  booktitle = {Proc. IEEE Int. Conf. on Robotics and Automation (ICRA)},
  pages     = {3699--3705},
  year      = {2022},
  doi       = {10.1109/ICRA46639.2022.9812193}
}

@inproceedings{sridhar2024nomad,
  author    = {Sridhar, Ajay and Shah, Dhruv and Glossop, Catherine and Levine, Sergey},
  title     = {NoMaD: Goal Masked Diffusion Policies for Navigation and Exploration},
  booktitle = {Proc. IEEE Int. Conf. on Robotics and Automation (ICRA)},
  pages     = {63--70},
  year      = {2024},
  doi       = {10.1109/ICRA57147.2024.10610665}
}

@inproceedings{chen2021waypoints,
  author    = {Chen, Changan and Majumder, Sagnik and Al-Halah, Ziad and
               Gao, Ruohan and Ramakrishnan, Santhosh Kumar and
               Grauman, Kristen},
  title     = {Learning to Set Waypoints for Audio-Visual Navigation},
  booktitle = {Proc. Int. Conf. on Learning Representations (ICLR)},
  year      = {2021},
  pages     ={1--16}
}

@inproceedings{yu2022soundadversarial,
  author    = {Yu, Yinfeng and Huang, Wenbing and Sun, Fuchun and
               Chen, Changan and Wang, Yikai and Liu, Xiaohong},
  title     = {Sound Adversarial Audio-Visual Navigation},
  booktitle = {Proc. Int. Conf. on Learning Representations (ICLR)},
  year      = {2022},
  pages     ={1--25}
}

@article{younes2023catch,
  author    = {Younes, Abdelrahman and Honerkamp, Daniel and
               Welschehold, Tim and Valada, Abhinav},
  title     = {Catch Me If You Hear Me: Audio-Visual Navigation in
               Complex Unmapped Environments with Moving Sounds},
  journal   = {IEEE Robotics and Automation Letters},
  year      = {2023},
  volume    = {8},
  number    = {2},
  pages     = {928--935},
  doi       = {10.1109/LRA.2023.3234766}
}

\end{document}